\pdfoutput=1

\documentclass[11pt]{article}

\usepackage{EACL2023}

\usepackage{times}
\usepackage{latexsym}
\usepackage{xcolor}    
\usepackage[most]{tcolorbox}

\usepackage[T1]{fontenc}

\usepackage[utf8]{inputenc}

\usepackage{microtype}

\usepackage{graphicx}
\graphicspath{ {./images/} }


\newcommand{\greybox}[0]{\includegraphics[width=.01\textwidth]{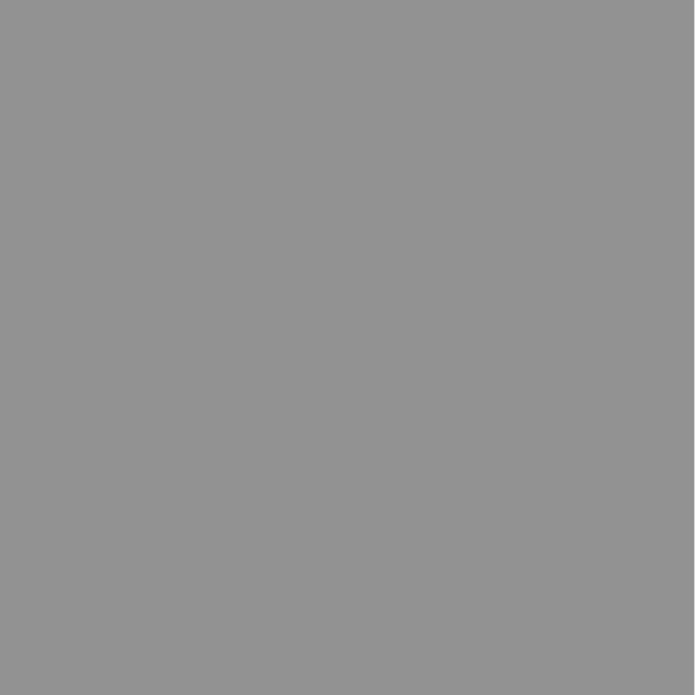}}
\newcommand{\redbox}[0]{\includegraphics[width=.01\textwidth]{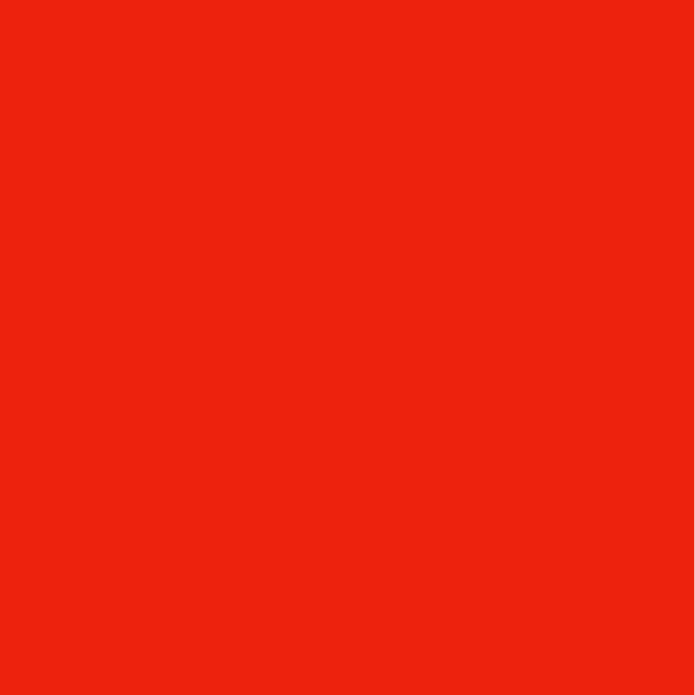}}
\usepackage{enumitem}

\usepackage[utf8]{inputenc}
\usepackage[english]{babel}
\usepackage{fancyhdr}
\usepackage{lastpage}
\usepackage{boldline}
\usepackage{caption}
\usepackage{subcaption}
\usepackage{cleveref}
\usepackage{tikz}
\usepackage{collcell}
\usepackage{xstring}
\usepackage{array}
\usepackage{colortbl}
\usepackage{numprint}
\usepackage{amssymb}
\usepackage{xcolor}
\usepackage{pifont}

\usepackage{arydshln}

\newcommand{\cmark}{\ding{51}}%
\newcommand{\xmark}{\ding{55}}%

\title{\textsc{AmbiCoref}:\\Evaluating Human and Model Sensitivity to Ambiguous Coreference}

\author{Yuewei Yuan \and
  Chaitanya Malaviya \and
  Mark Yatskar \\
  University of Pennsylvania \\
  \texttt{\{yuewei, cmalaviy, myatskar\}@seas.upenn.edu} \\}
  
\begin{document}
\maketitle
\begin{abstract}
Given a sentence ``Abby told Brittney that she upset Courtney'', one would struggle to understand who ``she'' refers to, and ask for clarification. However, if the word ``upset'' were replaced with ``hugged'', ``she'' unambiguously refers to Abby. We study if modern co-reference resolution models are sensitive to such pronominal ambiguity.
To this end, we construct \textbf{\textsc{AmbiCoref}}, a diagnostic corpus of minimal sentence pairs with ambiguous and unambiguous referents.
Our examples generalize psycholinguistic studies of human perception of ambiguity around particular arrangements of verbs and their arguments. 
Analysis shows that (1) humans are less sure of referents in ambiguous AmbiCoref examples than unambiguous ones, and (2) most coreference models show little difference in output between ambiguous and unambiguous pairs.
We release \textsc{AmbiCoref} as a diagnostic corpus for testing whether models treat ambiguity similarly to humans.\footnote{Our dataset and code is available at \href{https://github.com/LucyYYW/AmbiCoref}{https://github.com/LucyYYW/AmbiCoref.}}

\end{abstract}

\section{Introduction}
\begin{table*}[ht!]
\centering
\footnotesize	
\begin{tabular}{ |c| p{0.2\linewidth} | c | p{0.53\linewidth} | p{0.05\linewidth} | }
 \hlineB{6}
     & \textbf{Type} & \textbf{Ambig.}  & \textbf{Template} & \textbf{Count} 
     \\ \hlineB{0.01}
1 & Experiencer Obj (ECO-1)& \xmark & $[{\color{orange} Emily}]_A$ told $[Jessica]_B$ that $[\emph{\color{orange}she}]_A$ [saw] [Brian]. & 11336 \\ 
2 & Experiencer Obj (ECO-1)& \cmark & $[Emily]_A$ told $[Jessica]_B$ that $[\emph{she}]_?$ [bored] [Brian]. & 11336 \\ 
\hdashline
3 &Experiencer Obj (ECO-2)& \xmark & $[{\color{orange}The~mother}]_A$ told $[the~sister]_B$ that $[\emph{\color{orange}she}]_A$ [saw] the  client. &  11336 \\ 
4 &Experiencer Obj (ECO-2) & \cmark & $[The~mother]_A$ told $[the~sister]_B$ that $[\emph{she}]_?$ [bored] the client. & 11336 \\ 
\hlineB{0.01}

5 &Experiencer Sub (ECS-1) &\xmark & $[{\color{orange}The~aunt}]_A$ told $[Sarah]_B$ that [the~daughter] [met with] $[\emph{\color{orange}her}]_A$. & 4472 \\ 
6 &Experiencer Sub (ECS-1) &\cmark & $[The~aunt]_A$ told $[Sarah]_B$ that [the~daughter] [liked] $[\emph{her}]_?$. & 4472 \\ 
\hdashline
7 &Experiencer Sub (ECS-2)&\xmark & $[{\color{orange}The~father}]_A$ told $[the~son]_B$ that the client [met with] $[\emph{\color{orange}him}]_A$. & 4472 \\ 
8 &Experiencer Sub (ECS-2)&\cmark & $[The~father]_A$ told $[the~son]_B$ that the client [liked] $[\emph{him}]_?$. & 4472 \\ 

\hlineB{0.01}
9 &Implicit Causality (IC)&\xmark & $[{\color{orange}Abby}]_A$ [called] $[Jane]_B$ because $[\emph{\color{orange}she}]_A$ [wanted to apologize]. & 8424 \\ 
10 &Implicit Causality (IC)&\cmark & $[Abby]_A$ [called] $[Jane]_B$ because $[\emph{she}]_?$ [is leaving soon]. & 8424 \\ 
\hlineB{0.01}
11 &Transfer (TOP) &\xmark & $[Daniel]_A$ [baked] $[{\color{orange}the~boy}]_B$ [a cake] [after] $[\emph{{\color{orange}he}}]_B$ [asked for one]. & 8424 \\ 
12 &Transfer (TOP) &\cmark & $[Daniel]_A$ [baked] $[the~boy]_B$ [a cake] [before] $[\emph{he}]_?$ [had lunch]. & 8424 \\ 
 \hlineB{0.01}

\end{tabular}
\caption{Summary of the six template pairs that make up \textsc{AmbiCoref}. Template slot are indicated in square bracket, and clusters are marked with subscripts and color. All templates pair an unambiguous sentence with an ambiguous sentence, where they differ only in the choice of verb phrase.} 

\label{tab:templates}
\vspace{-10pt}
\end{table*}

\label{sec:introduction}



Ambiguity is a fundamental feature of language~\citep{puzzleOfAmbiguity} 
that some linguists believe
arises because of a pressure for efficient
communication \citep{haywood2005speakers,PIANTADOSI2012280}.
Recently, several works have highlighted the existence of ambiguity in tasks such as
question answering \citep{min-etal-2020-ambigqa,guo2021abg}, frame disambiguation \cite{dumitrache-etal-2019-crowdsourced}, anaphora resolution \citep{poesio-artstein-2005-reliability} and language modeling \cite{aina-linzen-2021-language}.
Yet systematic evaluation of how models react to ambiguity across many types of language processing problems is missing.
We contribute one such study about coreference resolution.

Coreference resolution is crucial to natural language understanding, especially in long contexts, such as dialog.
Ambiguity may arise naturally in dialog, but existing models do not have well-defined target behavior for such coreferences.
In contrast, when 
people encounter coreferential ambiguity, they recognize it, and can ask for clarification.
Existing resources, such as OntoNotes \cite{weischedel2013ontonotes}, do not provide fine-grained annotations of such instances to evaluate model behavior. This may result in models not being calibrated to handle the uncertainty in interpretations of ambiguous statements.
In this work, we ask how sensitive to ambiguity are models trained on these resources?


To understand how existing coreference models react to ambiguity, we construct a diagnostic corpus, \textbf{\textsc{AmbiCoref}}. 
\textsc{AmbiCoref} is composed of minimal pairs with ambiguous and unambiguous referents, created from four types of templates.
Ambiguity is achieved by reducing context sizes to one sentence, and creating sentences where participating verbs under-constrain the interpretation of their arguments.
For example, in Table~\ref{tab:templates}, line 2, our first template leverages ambiguity around verbs expressing subjective experiences.\footnote{Such instances require specific syntactic arrangements: the ambiguous instance in line 2 is unambiguous if the pronoun is moved to the object position of bored.}
The templates are designed by drawing on psycholinguistic studies~\cite{Springston,CARAMAZZA1977601,Rohde2014} and a core contribution of our work is to generalize their observations to create thousands of instances.
We achieve this by identifying VerbNet~\cite{verbnet} classes that are likely to contain appropriate verbs, and manually assigning them to templates.
Combined with variability we introduce using noun lists, \textsc{AmbiCoref} contains over 96 thousand sentences.

We verify that humans perceive instances in \textsc{AmbiCoref} in intended ways by crowdsourcing judgements (\S\ref{sec:human}). 
Annotators are asked to find the coreferent for a pronoun in a sentence, and rate their confidence, to account for the gradience in ambiguity judgements~\cite{schutze1995ambiguity}.
We find that, for unambiguous instances, humans strongly associate the pronoun with the intended noun but for ambiguous ones, they show reduced confidence across all templates, where the majority of participants are either not confident or mark them as ambiguous.
This suggests that humans process ambiguous and unambiguous sentences in \textsc{AmbiCoref} in qualitatively different ways.


\textsc{AmbiCoref} can be used to evaluate model behavior in the presence of ambiguity. 
We analyze five representative English models: three in CoreNLP~\cite{corenlp}, SpanBERT~\cite{coref_spanbert}, and NeuralCoref 4.0~\cite{wolf2019huggingface} (\S\ref{sec:evaluation}). 
Our main evaluation involves comparing coreference cluster assignments of the pronoun, between ambiguous and unambiguous samples.
4 out of the 5 models we analyze show almost no behavioral change. 
Unlike humans, coreference models largely do not alter their decisions in the presence of ambiguity.
Our analysis implies models likely need to explicitly account for ambiguity to achieve human-like behavior in the face of ambiguous input.

\section{Dataset Construction}
\begin{figure*}[ht!]
    \centering
    \includegraphics[width=.95\textwidth,trim={0 0 0 0}, clip]{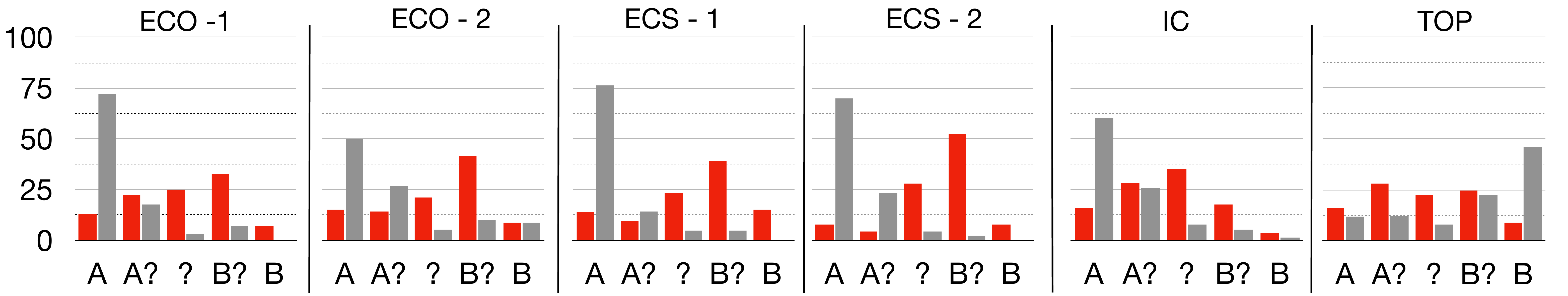}
    \caption{Human annotation of ambiguous (\redbox) and unambiguous (\greybox) sentences. We abbreviate human annotations by whether they identified noun A or B and whether they annotate definitely or likely (marked with ?). For example A? indicates, noun A, likely. The ground truth for unambiguous instances, from left to right, corresponds to A, A, A, A, A, B. Annotators read unambiguous examples as intended, and reduce their confidence on ambiguous examples.}
    \label{fig:human}
\end{figure*}
\label{sec:length}
To understand model sensitivity towards coreferential ambiguity, we build \textsc{AmbiCoref} using four types of templates, shown in Table \ref{tab:templates}.
The templates are created in minimal pairs, and the only difference between the ambiguous and unambiguous counterparts lies in the choice of verb phrase. Note that while ambiguity is a graded phenomenon, we use the the term ``ambiguous" for instances that are \textit{more likely} to elicit ambiguous human judgements and vice-versa.
Verb phrases are extracted from suitable verb classes in VerbNet~\cite{verbnet}, identified by manual annotation of VerbNet clusters.\footnote{We consider verbs from verb classes 31: Psych-Verbs (Verbs of Psychological State), 13: Verbs of Change of Possession, 37: Verbs of Communication as they conceptually align well with conditions required for ambiguity. Verbs within clusters were individually evaluated for appropriateness for templates by the authors.}
Each template is instantiated with verbs, names, noun-phrases, and gender-appropriate pronouns, greatly expanding the variation in cases identified in previous studies.

\subsection{Template Types}

\paragraph{Experiencer Constraint for Objects (ECO)}
\citet{Springston} propose the  Experiencer Constraint for complement constructions which we operationalize in our templates.
Verbs that mark their object as the experiencer of an emotion restrict the assignment of an object position pronoun to the subject of a declarative communication verb.
Conversely, the assignment is unconstrained when the pronoun is the subject of an experiencer verb.
For example, in row 2 of Table~\ref{tab:templates}, a pronoun in the subject position of ``bored'' is ambiguous (but would not be so in the object position). 
If the main verb does not impose an experiencer constraint, row 1, then a pronoun in the subject position is unambiguous. 
We instantiate two variants with names (rows 1,2) and general entities (rows 3,4).






\paragraph{Experiencer Constraint for Subjects (ECS)}
The Experiencer Constraint also suggests that verbs that mark their subjects as the experiencer of the emotion restrict the assignment of a subject position pronoun.
The assignment of the pronoun is unconstrained when it is in the object position.
For example, in Table~\ref{tab:templates}, row 6, ``liked'' is ambiguous when a pronoun is placed in the object position (but not in the subject position). 
We instantiate variants with names (rows 5,6) and entities (rows 7,8).

%


\paragraph{Implicit Causality (IC)} 
\citet{CARAMAZZA1977601} hypothesize that implicit causality of a verb can determine the direction of pronoun assignment.
For example, in Table~\ref{tab:templates} row 9, the phrase ``wanted to apologize'' establishes a cause for why ``Emily called,'' so the pronoun is constrained to the subject of ``call''.
Conversely, in row 10, the phrase ``is leaving soon'' fails to create such a relationship, leaving the pronoun ambiguous. 
For these templates (rows 9,10), we vary the names of the entities involved, and pair verbs (i.e. called) with constructed phrases that imply causality (i.e. apologizing), manually. 
\paragraph{Transfer of Possession (TOP)} 
\citet{Rohde2014} suggests that in transfer-of-possession contexts such as, ``John passed the comic to Bill. \emph{He}...'', the pronoun is equally likely to refer back to subject and non-subject.
We draw upon this observation, and create a template around verbs that involve source-goal possession transfers.
We distill the example to one sentence and pair the transfer event with a reason. 
For example, in Table~\ref{tab:templates} row 11, the phrase ``asked for one'' constrains the pronoun to be the receiver of ``bake''. 
Conversely, before having lunch provides no such constraint, because either the receiver or giver could have ``had lunch'' before the event.
Templates vary the names, verbs, objects, reasons, and preposition (rows 11,12).

\subsection{Filling Template Slots}
For each template, we construct a list of appropriate verb phrases, reasons (for IC and TOP templates), and shared list of gendered names and noun-phrases. 
Verb phrases were constructed by manually inspecting VerbNet classes.
To control for name bias, we randomly sample names from popular name lists\footnote{https://www.ssa.gov/oact/babynames/decades/} from the last 50 years, and reuse gendered noun-phrase lists from WinoBias~\cite{zhaoetal2018}. 
%
Excluding name and noun-phrase variations, templates have 114, 45, 81, 82 instances for ECO, ECS, IC, and TOP, respectively.






\section{Human Judgements}
\label{sec:human}

The templates used to create \textsc{AmbiCoref} generalize several psycholinguistic studies using lexical resources. 
Next, we verify that humans perceive ambiguity in these examples in the intended ways. 
We extract a subset of data for each template and ask Amazon Mechanical Turk workers which person a pronoun refers to (marked as \textit{A} or \textit{B} in Table~\ref{tab:templates}) and assign confidence (\textit{definitely}, or \textit{likely}). 
Annotators were also allowed to mark the referent as entirely \textit{ambiguous}. 
One sentence was sampled for each template and verb slot, uniformly at random. 
We collected 3 annotations per instance.\footnote{In ambiguous cases, annotators do not reliably annotate a particular category, but often guess with low confidence. As such, we do not only report a majority opinion per instance, but instead simply report multiple annotations per sentence to see overall trends.} See Appendix~\ref{app:mturk} for details on the collection of human judgements.

\begin{figure*}[t!]
    \centering
    \includegraphics[scale=0.23]{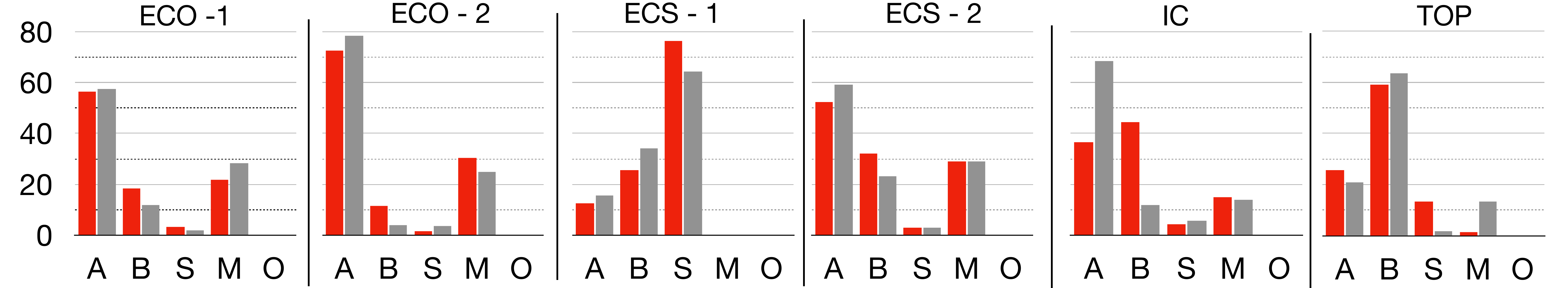}
    \caption{Percentage of ambiguous (\redbox) and unambiguous (\greybox)  instances that fall into each of our five cases for the SpanBERT-based model across all templates. All other models show negligible shifts (red and grey distributions are almost identical). The ground truth for unambiguous instances, from left to right, corresponds to A, A, A, A, A, B. 
    }
    \label{fig:spanbert_dist}
\end{figure*}

Figure~\ref{fig:human} summarizes our results.
Human judgments for unambiguous templates favor the intended coreference decision. 
For unambiguous ECO, ECS, IC, TOP instances, the intended reading is selected as likely or definitely, 83.2\%, 91.9\%, and 85.8\%, 68.3\% of the time, respectively.
For ambiguous instances, annotations display a substantial shift toward ambiguity. 
As shown in previous work, humans display substantial disagreement on ambiguous instances ~\cite{poesio-etal-2019-crowdsourced}. 
This is reflected in many templates, such as TOP, where humans produce almost uniform responses. 

\section{Model Evaluation}
\label{sec:evaluation}

We now examine if we can detect sensitivity to ambiguity in existing coreference resolution models by evaluating on \textsc{AmbiCoref}.
We experiment~\footnote{Roughly one week of continuous Colab GPU compute.} with five representative models: NeuralCoref 4.0 model from Hugging Face~\footnote{https://github.com/huggingface/neuralcoref}, SpanBERT~\cite{coref_spanbert} representation within the independent framework for end-to-end coreference~\cite{coref_e2e}, and the three models in Stanford CoreNLP~\cite{corenlp}:  deterministic~\cite{coref_deterministic}, statistical~\cite{coref_statistical} and neural mention ranking~\cite{coref_neural}. All models were trained on the CoNLL 2012 dataset \cite{pradhan2012conll}.

Here, we evaluate the model's final predictions, not their distribution over possible choices.
The reason is two-fold: (1) not all models produce a distribution and (2) initial analysis revealed that the models are miscalibrated, as in other settings \cite{desai-durrett-2020-calibration,jiang2021-how}, making it unreliable to interpret their output scores directly.

\subsection{Setup}
In this section, we ask, are there differences between how models process similar unambiguous and ambiguous examples?
As our examples are synthetically generated, we use the unambiguous examples as a form of control. 
If a model is unable to link the pronoun with the correct noun on unambiguous examples for at least 40\% of examples, we omit that template during evaluation.

We analyze model behavior by breaking it into cases that cover all possible cluster assignments for the pronoun in a single sentence. 
We compute the percentage of time a model outputs a cluster with:

\begin{itemize}[noitemsep]
\item case {\bf A}: the pronoun and noun A
\item case {\bf B}: the pronoun and noun B 
\item case {\bf S}: the pronoun as a singleton
\item case {\bf M}: the pronoun, noun A, and noun B 
\item case {\bf O}: the pronoun and any other span  
\end{itemize}

\begin{table}[t!]
\small
\begin{center}
\begin{tabular}{|c| c| c |}
\hline
Model &  Mean EMD \% & Templates \\
\hline
SpanBERT & 11.7 & 5\\
\hline
CoreNLP Neural & 3.5 & 5\\
\hline
NeuralCoref 4.0 & 4.0 & 5\\ 
\hline
CoreNLP Statistical & 1.2 & 3\\
\hline
CoreNLP Deterministic & 0.6 & 5 \\
\hline
\end{tabular}
\end{center}
\caption{Mean Earth Mover's Distance between matched ambiguous and unambiguous case distributions and the number of templates where models get at least 40\% of unambiguous cases correct. 
}
\label{tab:results}
\vspace{-10pt}
\end{table}

For example, Figure~\ref{fig:spanbert_dist} contains SpanBERT's output distribution over these cases for each template.
For each such distribution where the model's performance is above threshold, we compare ambiguous (red bar) and unambiguous (grey bar) distributions using Earth Mover's Distance (EMD) ~\cite{pele2009}\footnote{Earth Mover’s distances represent the amount of probability mass required to match two probability distributions. Hence, they help us compare distributions for ambiguous and unambiguous instances in a more interpretable way, than other possible measures like KL divergence.}. 
Table~\ref{tab:results} reports the number of templates above threshold, and their mean EMD. 

\subsection{Results}
Overall, most models we evaluated show essentially no change in output distribution over cases between ambiguous and unambiguous templates, as evidenced by near zero EMD. Most models are evaluated on five of six templates, but TOP is often excluded, representing a hard unambiguous case for most systems in its own right.

Of the models we evaluated, only SpanBERT shows significant deviation in behavior with ambiguous inputs.
Figure~\ref{fig:spanbert_dist} breaks down SpanBERT's performance on each template. 
While average EMD is higher than for other models, it still largely doesn't change predictions.
When decisions change, often the pronoun is linked with the other noun. 
For example, in ambiguous cases of ECO-1, SpanBERT reduces merged outputs, and instead links the pronoun with noun B more frequently.
In ambiguous cases, other models largely link the first noun-phrase (A) to the pronoun.

\section{Discussion and Conclusion}

Overall, our results suggest that model behavior significantly deviates from how human treat ambiguous coreference. 
We lend more evidence that models miss aspects of how people understand language, especially in discourse~\cite{upadhye-etal-2020-predicting}.
The reason is likely in part that models are trained on resources which do not account for distributions in judgments. As a result, models do not have well-defined behavior when ambiguity arises and are poorly calibrated.

Training models with finer-grained coreference judgments could allow models to better align with human behavior. Techniques to improve model calibration could also be effective, allowing models to abstain or seek clarification when ambiguity arises.
We hope that \textsc{AmbiCoref} can serve as a diagnostic set for future modeling approaches in evaluating their sensitivity to instances of ambiguity in language.
\section{Limitations}
Our study focuses entirely on coreference in the English language with models trained in high-resource settings. 
Furthermore, the cases of ambiguity we identify are English-specific and the names we insert into templates are popular American names.
It is an open question as to how our results generalize to low-resource non-American-English settings. 

The language we use to evaluate models is templatic. 
While we make an effort to account for unnatural data, by only evaluating templates models do well at, models struggle to completely solve all our unambiguous examples. 
This presents a challenge for future model builders.
On the other hand, our templates may not reflect a particular real world distribution that models will be tested on.

\section*{Acknowledgements}

We thank Chris Callison-Burch and the PennNLP group for their helpful comments on this work.


\bibliography{references}
\bibliographystyle{acl_natbib}

\clearpage
\appendix
\section{Human Judgement Tests}
\label{app:mturk}

\begin{figure*}
    \centering
    \includegraphics[width=\textwidth,keepaspectratio=True]{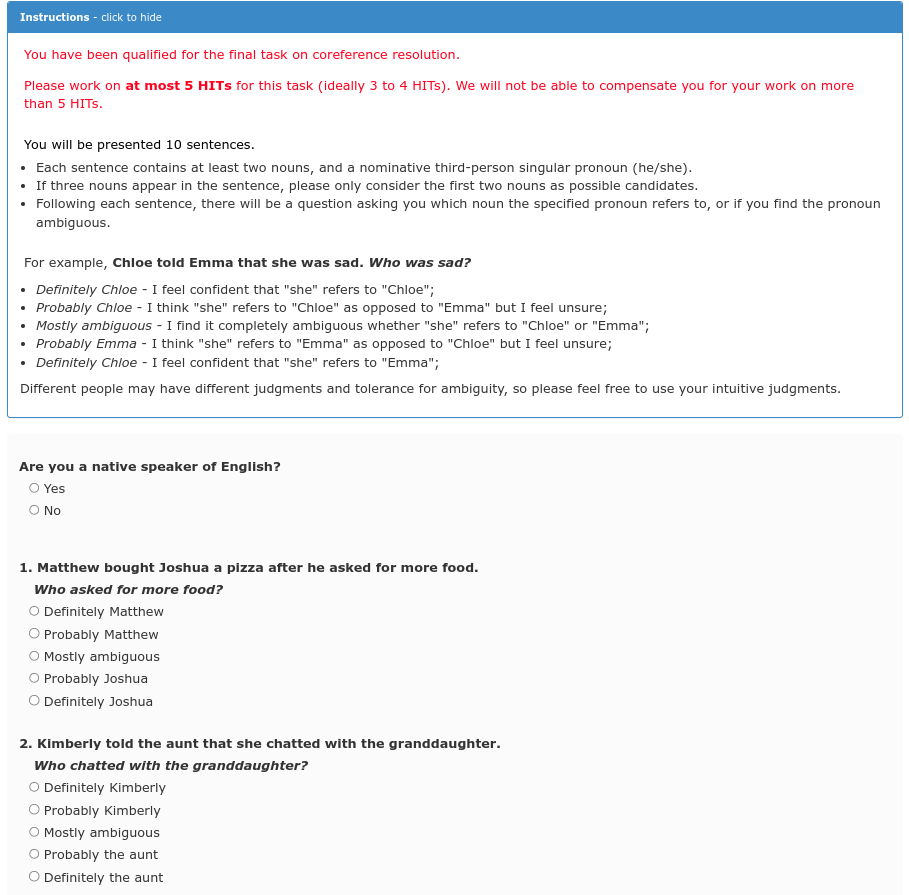}
    \caption{Annotation interface for the human judgement tests, presented in section~\ref{sec:length}.}
    \label{fig:interface}
\end{figure*}

In all our human judgement tests, we required annotators to be based primarily in English-speaking countries: the US, UK, Canada or Australia. Further, annotators needed to have at least 1000 approved HITs and a HIT acceptance rate of at least 98\%. Each HIT contained 10 examples, and we estimated the completion time for each HIT to be $\sim$5 minutes, so we paid \$1.25 per HIT, for a pay rate of \$15 per hour.

For our human judgement tests, we first ran a qualification round to ensure high-quality annotations. In this round, we asked annotators to complete a single HIT with 10 examples (5 unambiguous, 5 ambiguous randomly ordered). For each annotator who completed this round, we compute their accuracy by measuring how often they responded with the correct referent (or the ambiguous label), while ignoring their confidence. The top 100 annotators were qualified to work on the main task.

For our main task, we had 625 sentences labeled in total, with 3 assignments per sentence. Each annotator was asked to work on not more than 5 HITs, so that we get a diverse set of judgements.
Similar to the qualification round, we asked each annotator to label the referent (or the ambiguous label) and their confidence. We group the annotations into 5 options: (Noun A, definitely), (Noun A, likely), Ambiguous, (Noun B, likely), and (Noun B, definitely). The human judgement labels for each template type were aggregated by computing the fraction of annotations in each of the five options. 
Our annotation interface for the main task is shown in Figure~\ref{fig:interface}.

\end{document}